\documentclass[letterpaper, 10 pt, conference]{ieeeconf}

\IEEEoverridecommandlockouts
\usepackage{amsmath,amssymb,bm,booktabs,graphicx,xcolor,url,multirow}
\usepackage{algorithm}
\usepackage[noend]{algpseudocode}
\usepackage{tikz}
\usetikzlibrary{arrows.meta,positioning,fit,backgrounds}

\usepackage{microtype}

\newcommand{\R}{\mathbb{R}}
\newcommand{\SEthree}{\mathrm{SE}(3)}
\newcommand{\sethree}{\mathfrak{se}(3)}
\newcommand{\vmin}{\bm{v}_{\min}}
\newcommand{\Om}{\bm{\Omega}}
\newcommand{\M}{\bm{M}}
\newcommand{\n}{\bm{n}}
\newcommand{\J}{\bm{J}}
\newcommand{\Hess}{\bm{H}}

\title{\LARGE \bf
LF-GICP: Parameter-Free Degeneracy-Aware LiDAR Odometry\\
via a Voxel-Normal Localizability Field}
\author{Eunsoo Im}

\begin{document}

\maketitle
\thispagestyle{empty}
\pagestyle{empty}

\begin{abstract}
Scan-to-map LiDAR odometry drifts unboundedly along the unobservable axes of
geometrically degenerate environments like tunnels and corridors, and
existing degeneracy handling requires environment-specific parameter tuning.
This paper presents a parameter-free approach. We show that in voxelized
GICP the Gauss--Newton (GN) Hessian masks translational degeneracy, because
covariance regularization keeps the translation block artificially
well-conditioned. We bypass this with a regularization-free
voxel-normal localizability field and two of its statistics: a
normalized fraction $f_0$ detecting directional anisotropy, and an absolute
per-voxel mass $\lambda_0$ distinguishing information absence
(tunnels) from dilution (dense open scenes). A temporal-median gate
combines both to trigger Fisher-information correspondence weighting.
Calibrated once by fixed rules on two short sequences and then frozen,
LF-GICP achieves the lowest KITTI relative translation error
($0.865\%$) under an identical evaluation protocol against re-run
baselines, outperforms them on GEODE tunnels and MulRan, leads the HeLiPR
mean, and generalizes across four sensor types without re-tuning.
We further demonstrate empirically that straight, uniform tunnels remain
unobservable along their axis for LiDAR-only registration.
\end{abstract}

\noindent\textit{Code:}~\url{https://github.com/ies0411/LF_GICP}

\section{Introduction}\label{sec:intro}

\begin{figure}[t]\centering
\includegraphics[width=\columnwidth]{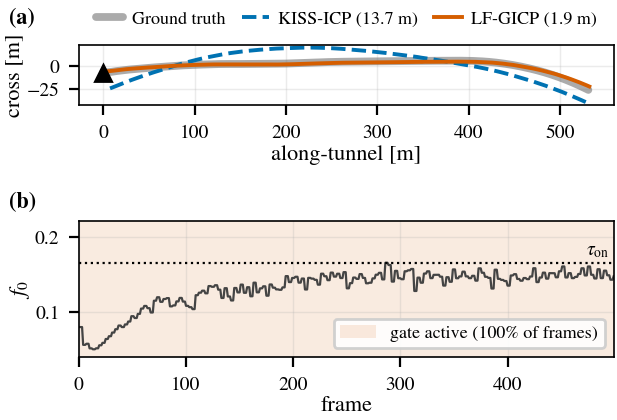}
\caption{GEODE Urban\_Tunnel02 under the \emph{same frozen configuration}
as the KITTI evaluation (Sec.~\ref{sec:kitti}). (a)~Aligned trajectories
(500-frame GT horizon): KISS-ICP drifts laterally while LF-GICP tracks
ground truth. (b)~$f_0$: the median gate detects the tunnel without an
environment flag or site-specific threshold.}
\label{fig:teaser}
\end{figure}

Well-engineered frame-to-map ICP pipelines such as KISS-ICP~\cite{kiss} and
the GICP family~\cite{gicp,vgicp} now rival complex LiDAR odometry systems
by generalizing across sensors with frozen parameters or per-voxel
statistics. Yet they fail under geometric degeneracy: in tunnels,
corridors, or mines the observed surfaces only partially constrain the
motion, the cost becomes flat along the unobservable direction, and drift
grows unboundedly (Fig.~\ref{fig:teaser})---a failure mode that has
motivated a decade of degeneracy-aware registration
research~\cite{zhangdeg,xicp,genz,lpicp}.


Existing degeneracy handling methods typically threshold optimization metrics, such as GN-Hessian eigenvalues~\cite{zhangdeg} or constraint-space alignments~\cite{xicp,lpicp}. Although they mitigate drift using priors~\cite{zhangdeg,xicp} or residual re-balancing~\cite{genz}, they invariably retain environment-dependent parameters, such as eigenvalue thresholds, dataset presets, and environment flags, which prevent out-of-the-box deployment. This paper introduces a parameter-free alternative, namely a single configuration calibrated once that operates across diverse datasets, sensors, and platforms. This approach rests on two key insights.

First, the GN Hessian of a voxelized GICP pipeline structurally masks
translational degeneracy. Point-to-distribution registration uses full voxel
covariances, so together with standard covariance regularization every
correspondence contributes an isotropic information floor to the Hessian's
translation block: empirically, the eigenvalue ratio
$\lambda_{\min}/\lambda_{\max}$ inside a uniform tunnel ($0.87$) is
indistinguishable from open-road driving ($0.81$). The floor is intrinsic to
the cost formulation, rendering Hessian-eigenvalue thresholding ill-posed in
GICP-style pipelines (Sec.~\ref{sec:hidden}). We therefore propose a
regularization-free \emph{voxel-normal localizability field}
$\M=\sum_v\rho_v\n_v\n_v^\top$ built from planarity-weighted map normals:
its normalized smallest eigenvalue $f_0$ isolates degenerate scenes and its
null eigenvector identifies the unobservable axis (Sec.~\ref{sec:field}).

Second, anisotropy alone does not imply degeneracy: information
dilution must be distinguished from information absence. A
uniform tunnel and a dense tree-lined boulevard produce similarly low $f_0$,
yet reweighting the weak axis helps in the former and hurts in the latter,
where many weakly informative points collectively constrain the estimate. We
separate the regimes with the absolute weak-axis normal mass per voxel,
$\lambda_0$ (Sec.~\ref{sec:lam0}): across five datasets and four sensor types
it cleanly bifurcates tunnels/mines ($0.005$--$0.013$) from open/vegetated
scenes ($0.017$--$0.033$). The mitigation gate enforces both conditions, with
thresholds derived by a fixed rule from two short calibration traces and
frozen thereafter.

\emph{Contributions.}
(1)~A structural diagnosis showing that the GN Hessian of
point-to-distribution registration hides translational degeneracy
(Sec.~\ref{sec:hidden}).
(2)~A regularization-free voxel-normal localizability field whose fraction
statistic $f_0$ isolates anisotropy and whose null eigenvector recovers the
unobservable axis (Sec.~\ref{sec:field}).
(3)~The absolute-mass statistic $\lambda_0$ distinguishing information
absence from dilution (Sec.~\ref{sec:lam0}).
(4)~A parameter-free protocol deriving all constants from two calibration
traces, validated across five benchmarks and four sensor types
(Secs.~\ref{sec:params},~\ref{sec:exp}).
(5)~An empirical analysis of the failure modes of alternative strategies,
mapping the boundaries of degeneracy-aware weighting
(Secs.~\ref{sec:limit}, \ref{sec:abl}).

\section{Related Work}\label{sec:related}

\textbf{LiDAR odometry.} Feature-based LOAM~\cite{loam} has largely given
way to direct scan-to-map pipelines built on classical
ICP~\cite{besl,chenmedioni}, such as KISS-ICP~\cite{kiss},
CT-ICP~\cite{ctcip}, DLO~\cite{dlo}, and MAD-ICP~\cite{madicp}. The
distribution-based family (NDT~\cite{ndt}, GICP~\cite{gicp,smallgicp})
models local surface statistics; the voxelized variant VGICP~\cite{vgicp}
is our backend. GenZ-ICP~\cite{genz} adaptively blends point-to-point and
point-to-plane residuals, a strong degeneracy-robust baseline. LiDAR-inertial systems~\cite{fastlio2,pointlio}
mitigate short-horizon degeneracy with IMUs. We focus on the LiDAR-only
problem, which matters when inertial data is unreliable and because
unmitigated degeneracy contaminates downstream fusion.

\textbf{Degeneracy detection.} ICP normal-equation conditioning has been
analyzed since stable sampling~\cite{gelfand} and closed-form
covariance~\cite{censi}; classical degeneracy handling thresholds
GN-Hessian eigenvalues~\cite{zhangdeg}, and newer approaches classify
directional localizability via constraint-space alignments
(X-ICP~\cite{xicp}, LP-ICP~\cite{lpicp}), environment-driven
estimation~\cite{zhen}, or learning~\cite{nubert}. A recent field study~\cite{ica} found that these
generalize poorly without site-specific tuning. As Sec.~\ref{sec:hidden}
shows, metrics computed in the optimization's information space are masked by
covariance regularization in point-to-distribution pipelines. In contrast,
our method directly evaluates the unregularized voxel-normal field. Filter-level and graph-level
approaches~\cite{hinduja,lodestar,dammloam} address error propagation and
are complementary.

\textbf{Degeneracy mitigation.} Prior strategies (i)~lock degenerate
directions to a prior~\cite{zhangdeg,xicp}, (ii)~reweight
residuals~\cite{genz}, or (iii)~augment with photometric or inertial
channels~\cite{coinlio,fastlio2}. Our analysis exposes limits of each: hard
constraints discard the weak but valid along-axis signal that GICP's
regularized solver retains (Sec.~\ref{sec:mit}); reweighting helps under
information absence but hurts under dilution (Sec.~\ref{sec:lam0}); and
photometry cannot resolve the null space of uniform tunnels
(Sec.~\ref{sec:limit}).

\textbf{Parameter-freeness and evaluation practice.} KISS-ICP~\cite{kiss}
established the single-frozen-configuration paradigm but lacks degeneracy
handling; GenZ-ICP~\cite{genz} shares the goal yet remains vulnerable in
tunnels (Table~\ref{tab:geode}). Degeneracy-aware systems typically retain
heuristic or site-level thresholds~\cite{xicp,lpicp} and are mainly
evaluated in degenerate scenes; a field study documents their
sensitivity~\cite{ica}. LF-GICP is the first to hold a single configuration
across open-road SOTA benchmarks and severe tunnels~\cite{kitti,mulran,helipr,geode};
all gate constants come from calibration rules on two short traces
(Sec.~\ref{sec:params}), backed by a sensitivity analysis
(Sec.~\ref{sec:abl}).


\section{Method}\label{sec:method}

Fig.~\ref{fig:pipeline} overviews LF-GICP. We outline the baseline
optimization (Sec.~\ref{sec:form}), analyze why Hessian-based metrics
structurally fail in point-to-distribution pipelines (Sec.~\ref{sec:hidden}),
construct the localizability field and its two statistics
(Secs.~\ref{sec:field},~\ref{sec:lam0}), and detail the weighting, gating,
and LiDAR-only estimation limits (Secs.~\ref{sec:mit}--\ref{sec:limit}).

\definecolor{lfor}{HTML}{D55E00}
\definecolor{lfbl}{HTML}{0072B2}
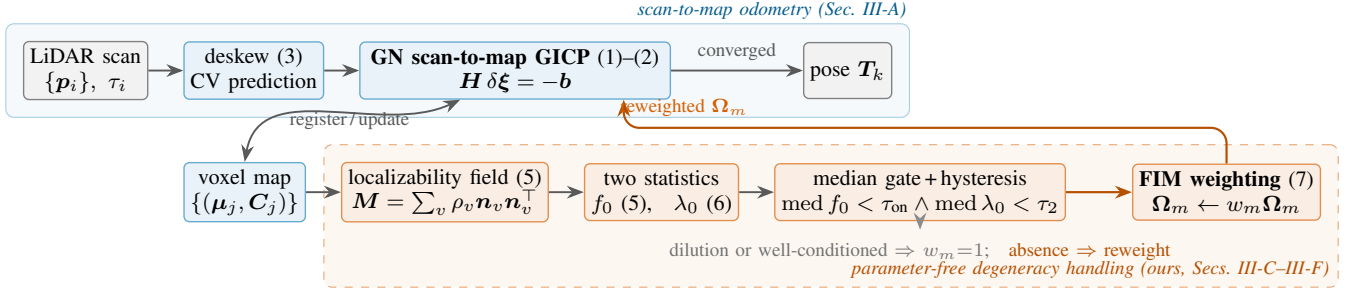
\begin{figure*}[t]\centering
\resizebox{\textwidth}{!}{%
\begin{tikzpicture}[
  font=\footnotesize,
  node distance=4mm and 4.5mm,
  box/.style={draw=black!55, rounded corners=2pt, line width=0.5pt,
              minimum height=7.2mm, inner sep=2.4pt, align=center, fill=white},
  base/.style={box, draw=lfbl!60, fill=lfbl!9},
  ours/.style={box, draw=lfor!70, fill=lfor!11},
  io/.style={box, draw=black!50, fill=black!5},
  arr/.style={-{Stealth[length=2.2mm]}, line width=0.7pt, black!65},
  arro/.style={-{Stealth[length=2.2mm]}, line width=0.8pt, lfor!85!black},
]
\node[io] (scan) {LiDAR scan\\$\{\bm p_i\},\ \tau_i$};
\node[base, right=of scan] (deskew) {deskew~\eqref{eq:deskew}\\CV prediction};
\node[base, right=of deskew, minimum width=40mm] (gn)
  {\textbf{GN scan-to-map GICP}~\eqref{eq:cost}--\eqref{eq:gn}\\$\Hess\,\delta\bm\xi=-\bm b$};
\node[io, right=17mm of gn] (pose) {pose $\bm T_k$};
\node[base, below=8mm of deskew.south west, anchor=north west] (map)
  {voxel map\\$\{(\bm\mu_j,\bm C_j)\}$};
\node[ours, right=of map] (field)
  {localizability field~\eqref{eq:M}\\$\M=\sum_v \rho_v \n_v\n_v^\top$};
\node[ours, right=of field] (stats)
  {two statistics\\$f_0$~\eqref{eq:M},\quad$\lambda_0$~\eqref{eq:lam0}};
\node[ours, right=of stats] (gate)
  {median gate\,+\,hysteresis\\$\mathrm{med}\,f_0<\tau_{\text{on}} \wedge \mathrm{med}\,\lambda_0<\tau_2$};
\node[ours, right=8.5mm of gate] (fim)
  {\textbf{FIM weighting}~\eqref{eq:c1}\\$\Om_m\leftarrow w_m\Om_m$};
\node[font=\scriptsize, text=black!55, below=1.6mm of gate, align=center] (dil)
  {dilution or well-conditioned $\Rightarrow$ $w_m{=}1$;\quad
   \textcolor{lfor!85!black}{absence $\Rightarrow$ reweight}};
\draw[arr] (scan) -- (deskew);
\draw[arr] (deskew) -- (gn);
\draw[arr] (gn) -- node[above]{\scriptsize converged} (pose);
\draw[{Stealth[length=2.2mm]}-{Stealth[length=2.2mm]}, line width=0.7pt, black!65]
  (map.north) to[out=80,in=205]
  node[right=0.8mm, pos=0.25]{\scriptsize register\,/\,update} (gn.208);
\draw[arr] (map) -- (field);
\draw[arr] (field) -- (stats);
\draw[arr] (stats) -- (gate);
\draw[arro] (gate) -- (fim);
\draw[arr, black!45, shorten >=0.5mm] (gate.south) -- (dil.north);
\draw[arro, rounded corners=2mm] (fim.north) -- ++(0,4.6mm) -|
  node[pos=0.45, above=0.2mm]{\scriptsize\textcolor{lfor}{reweighted $\Om_m$}}
  ([xshift=14mm]gn.south);
\begin{scope}[on background layer]
\node[fill=lfbl!4, draw=lfbl!35, rounded corners=3pt, inner sep=2.2mm,
      fit=(scan)(deskew)(gn)(pose)] (band1) {};
\node[anchor=south east, font=\scriptsize\itshape, text=lfbl!80!black,
  inner sep=1.2pt] at (band1.north east)
  {scan-to-map odometry (Sec.~\ref{sec:form})};
\node[fill=lfor!5, draw=lfor!55, dashed, rounded corners=3pt, inner sep=2.2mm,
      fit=(field)(stats)(gate)(fim)(dil)] (band2) {};
\node[anchor=south east, font=\scriptsize\itshape, text=lfor!85!black]
  at (band2.south east) {parameter-free degeneracy handling (ours,
  Secs.~\ref{sec:field}--\ref{sec:gate})};
\end{scope}
\end{tikzpicture}%
}
\caption{LF-GICP overview. From the accumulated voxel
map---\emph{before} covariance regularization---we build the localizability
field $\M$ and its two statistics: the anisotropy fraction $f_0$ and the
absolute weak-axis mass $\lambda_0$ that separates information
\emph{absence} from \emph{dilution}. A trailing-median gate with hysteresis
activates Fisher-information weighting only under genuine absence; diluted
or well-conditioned scenes run the loop (top, blue) untouched.}
\label{fig:pipeline}
\end{figure*}

\subsection{Scan-to-Map Registration Baseline}\label{sec:form}
Let the source scan be represented by points $\{\bm p_i\in\R^3\}_{i=1}^{N_s}$ and the target as a voxel map of Gaussians $\{(\bm\mu_j,\bm C_j)\}_{j=1}^{N}$, where $\bm\mu_j$ and $\bm C_j$ denote the mean and covariance of the points within voxel $j$. Generalized-ICP estimates the rigid body pose $\bm T=(\bm R,\bm t)\in\SEthree$ by minimizing the Mahalanobis point-to-distribution cost:
\begin{equation}
\bm T^\star=\arg\min_{\bm T}\sum_{m}\bm d_m^\top\Om_m\bm d_m,\quad
\bm d_m=\bm R\bm p_{i(m)}+\bm t-\bm\mu_{j(m)},
\label{eq:cost}
\end{equation}
with per-correspondence information $\Om_m=\bm C_{j(m)}^{-1}\in\R^{3\times3}$. Linearizing $\bm d_m$ with respect to the local twist $\delta\bm\xi=[\delta\bm\omega;\delta\bm\rho]\in\R^6$ (rotation; translation) yields the Gauss--Newton (GN) normal equations:
\begin{equation}
\Hess\,\delta\bm\xi=-\bm b,\quad
\Hess=\sum_m\J_m^\top\Om_m\J_m,\quad
\bm b=\sum_m\J_m^\top\Om_m\bm d_m,
\label{eq:gn}
\end{equation}
where $\J_m=\big[-[\bm R\bm p_{i(m)}+\bm t]_\times \quad \bm I_3\big]\in\R^{3\times6}$. The Hessian $\Hess$ is the empirical Fisher Information Matrix (FIM); the rank deficiency of its translation block $\Hess_{\rho\rho}=\sum_m\Om_m$ is the estimation-theoretic signature of translational degeneracy. Updates $\bm T\leftarrow\exp(\delta\bm\xi)\,\bm T$ iterate until convergence.

To ensure that the proposed degeneracy handling is the sole evaluation
variable, the baseline adopts a standard configuration: constant-velocity
(CV) initial predictions, a KISS-style adaptive correspondence threshold, a
Huber weight ($\delta{=}1$) composed multiplicatively with the FIM weight of
Sec.~\ref{sec:mit}, sensor-class voxel sizes ($1.0$\,m for $\ge$64 beams,
$0.5$\,m for $\le$32), and an age-windowed local map of per-voxel Gaussians
($\ge$3 points, last $500$ scans) with re-association at each GN iteration.
Because point-to-distribution GICP is highly sensitive to spinning-LiDAR
intra-scan distortion ($\sim$1--1.5\,m), we deskew whenever per-point
timestamps $\tau_i\in[0,1]$ are available (reconstructed via azimuth for
MulRan), transforming each point into the scan-end frame via the closed-form
$\SEthree$ vectorization
\begin{equation}
\bm p_i\leftarrow\exp\!\big((\tau_i-1)\,\bm\xi\big)\,\bm p_i ,
\label{eq:deskew}
\end{equation}
where $\bm\xi=\log(\bm T_{k-2}^{-1}\bm T_{k-1})\in\sethree$ is the body-frame
twist of the previous scan under constant velocity.

\subsection{Structural Masking in the GN Hessian}\label{sec:hidden}
Covariance regularization acts as $\bm C_j \leftarrow \bm C_j + \beta \bm I$,
and information follows by inversion. A planar voxel with normal $\n_j$ has a
small normal variance $\sigma_n^2$ and a large in-plane variance
$\sigma_t^2$; its regularized covariance inverts exactly to
$\Om_j=\kappa_t\bm I+(\kappa_n-\kappa_t)\,\n_j\n_j^\top$
with $\kappa_n=1/(\sigma_n^2+\beta)\gg\kappa_t=1/(\sigma_t^2+\beta)$. The
translation block of \eqref{eq:gn} is therefore
\begin{equation}
\Hess_{\rho\rho}=\sum_m\Om_m
=\underbrace{\Big(\sum_m\kappa_t^{(m)}\Big)\bm I}_{\text{isotropic floor}}
+\sum_m(\kappa_n^{(m)}-\kappa_t^{(m)})\,\n_m\n_m^\top .
\label{eq:floor}
\end{equation}
Since the floor scales linearly with the number of correspondences, writing
$\Hess_{\rho\rho}=F\bm I+\bm S$ with $F=\sum_m\kappa_t^{(m)}$ and
$\bm S=\sum_m(\kappa_n^{(m)}-\kappa_t^{(m)})\,\n_m\n_m^\top\succeq0$ yields
$\lambda_{\min}/\lambda_{\max}=(F+s_1)/(F+s_3)\ge\bar\kappa_t/\bar\kappa_n$,
a bound set only by the material statistics
$(\sigma_n^2,\sigma_t^2,\beta)$, regardless of scene geometry. Crucially, the
floor does not vanish as $\beta\to0$: $\kappa_t\to1/\sigma_t^2$ stays finite
because the in-plane variance is \emph{intrinsic} to the full-covariance
formulation---every correspondence contributes information in all three
dimensions, not only along its normal.

Real-world data confirms this masking. In the degenerate GEODE Metro shield
tunnel, $\Hess_{\rho\rho}$ yields $\lambda_{\min}/\lambda_{\max}=0.87$ and
$\lambda_{\min}/\mathrm{tr}=0.31$, statistically identical to KITTI open-road
driving ($0.81$ and $0.29$). Nor is this a tunable artifact: lowering $\beta$
from $2.0$ to $0.1$ leaves the tunnel ratio at $0.32$ while degrading
KITTI-00 ATE from $0.318$ to $0.401$\,m (Table~\ref{tab:creg}).
Hessian-eigenvalue detectors therefore inherently require
environment-dependent tuning, explaining the site-specific parameter
reliance of prior work.

\subsection{Voxel-normal localizability field}\label{sec:field}
To bypass the masking, we detect degeneracy from the raw geometry
\emph{before} regularization. For each local-map voxel $v$, the normal
$\n_v$ is the dominant eigenvector of $\Om_v$, with planarity
$\rho_v=(\lambda_3-\lambda_2)/\lambda_3\in[0,1]$
($\lambda_1\!\le\!\lambda_2\!\le\!\lambda_3$: eigenvalues of $\Om_v$); for a
planar voxel $\rho_v=(\kappa_n-\kappa_t)/\kappa_n$ approaches $1$, and $0$
for an isotropic one. The \emph{localizability field} $\M$ and its ratio
metric $f_0$ are:
\begin{equation}
\M=\sum_{v}\rho_v\,\n_v\n_v^\top\in\R^{3\times3},\qquad
f_0=\frac{\lambda_{\min}(\M)}{\mathrm{tr}(\M)}\in[0,\tfrac13].
\label{eq:M}
\end{equation}

Because $\M$ accumulates only planarity-weighted normals, it carries no
isotropic floor. In degenerate corridors the normals span only the
cross-section plane, rendering $f_0\!\to\!0$, while the minimal eigenvector
$\bm u=\arg\min_{\|\bm x\|=1}\bm x^\top\M\bm x$ identifies the unobservable
translation axis. Critically, if all contributing voxels are ideal planes
with common normal information $\kappa_n$, the unregularized translation FIM
equals $\kappa_n\M$ up to the intrinsic in-plane floor, so
$\mathrm{null}(\M)$ defines the unconstrained translation subspace; the
isotropic floor in \eqref{eq:floor} shifts the eigenvalues of
$\Hess_{\rho\rho}$ but leaves $\M$ itself unchanged. The statistic is
scale-free and does not saturate as the map expands; stride-subsampling to
$L_{\max}\!=\!4096$ voxels keeps the once-per-frame evaluation at
microsecond cost independent of map size.

\subsection{Information Absence and Dilution Classification}\label{sec:lam0}
Because $f_0$ is a scale-free ratio, open scenes with merely imbalanced
normals (flat highways, boulevards) can score as low as a tunnel where
weak-axis information is genuinely absent. The distinction is vital:
concentrating weights on the weak axis is effective under absence but
degrades accuracy by up to 1.4$\times$ under dilution
(Sec.~\ref{sec:abl}). We therefore introduce a second, regularization-free
statistic, the absolute weak-axis normal mass per sampled voxel:
\begin{equation}
\lambda_0=\frac{\lambda_{\min}(\M)}{|\mathcal V|},
\label{eq:lam0}
\end{equation}
where $|\mathcal V|\le L_{\max}$ is the number of sampled voxels.

\begin{figure}[t]\centering
\includegraphics[width=\columnwidth]{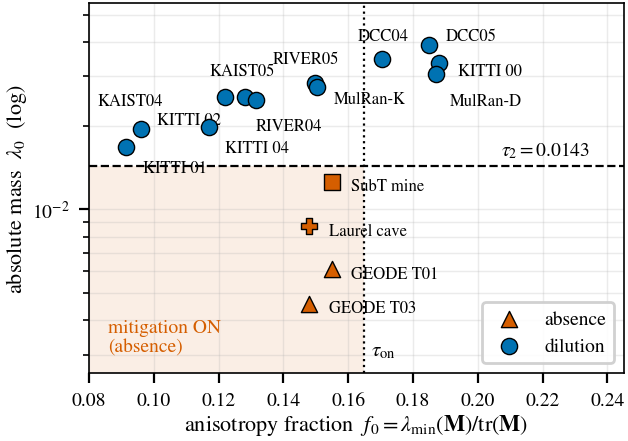}
\caption{Sequence medians of $(f_0,\lambda_0)$ across five benchmarks and
four sensor types. $f_0$ alone cannot separate tunnels (absence) from
anisotropic but data-rich scenes (dilution); $\lambda_0$ bifurcates the
classes with a wide margin around the calibrated $\tau_2$. Mitigation
triggers only in the shaded region.}
\label{fig:scatter}
\end{figure}

As illustrated in Fig.~\ref{fig:scatter}, $\lambda_0$ cleanly separates information absence (GEODE, Laurel, SubT spanning 0.0046--0.0126) from dilution (KITTI, HeLiPR, MulRan spanning 0.0168--0.0334).

The metric generalizes across 16-, 64-, and 128-channel sensors because
voxel resolution scales with the sensor class. Analytically, if the
weak-axis normal count is a fraction $\epsilon$ of the cross-section count,
then $f_0 = \epsilon/(1+\epsilon)$ regardless of scene density, whereas
$\lambda_0 \approx \epsilon\bar\rho$ scales with the actual weak-axis surface
mass ($\bar\rho$: mean planarity). Tunnels drive $\epsilon \to 0$; dense
boulevards merely keep $\epsilon$ small while retaining large absolute mass.
Both depress the fraction, but only true absence depresses the mass---a
property of the map's normal field that per-correspondence weight histograms
cannot capture. Enforcing both low $f_0$ and low $\lambda_0$ therefore
restricts the reweighting of Sec.~\ref{sec:mit} to genuine absence.

\subsection{Soft Fisher Information Weighting}\label{sec:mit}
When degeneracy is present, we weight each correspondence by its
information along the weakest direction $\bm v_{\min}\in\R^6$, the minimal
eigenvector of the full GN Hessian $\Hess$ in \eqref{eq:gn}:
\begin{equation}
w_m=\frac{\bm v_{\min}^\top\J_m^\top\Om_m\J_m\,\bm v_{\min}}{\tfrac1{N_c}\sum_{m'}\bm v_{\min}^\top\J_{m'}^\top\Om_{m'}\J_{m'}\,\bm v_{\min}},\qquad \Om_m\leftarrow w_m\Om_m,
\label{eq:c1}
\end{equation}
after which \eqref{eq:gn} is re-solved ($N_c$: number of correspondences).
This focuses the update on the minority of correspondences that inform the
weak axis, at the cost of one $6\times6$ eigendecomposition per GN iteration
and one quadratic form per correspondence. Mean normalization preserves the
information scale and the adaptive-threshold statistics. With
$q_m=\bm v_{\min}^\top\J_m^\top\Om_m\J_m\bm v_{\min}$ and $w_m=q_m/\bar q$,
the quadratic-mean inequality gives
$\bm v_{\min}^\top\Hess'\bm v_{\min}=\sum_m q_m^2/\bar q\ge\sum_m q_m$, so
reweighting never reduces weak-axis information; each term of $\Hess'$ is a
nonnegative multiple of an existing measurement, so nothing is injected into
the data null space.

We avoid hard equality constraints that lock the degenerate axis to a motion
prior: GICP's regularized solver retains a weak but valid along-axis signal
that outperforms dead-reckoning, and replacing it with a hard lock to a CV
prior (similar to X-ICP) causes GEODE Urban to diverge and worsens GEODE
Metro. Soft weighting preserves this residual structural signal while
stabilizing the linear system.

\subsection{Temporal Gating and Operational Control Policy}\label{sec:gate}
Because raw per-frame statistics are noisy under transient occlusions, we
gate the mitigation state on trailing-window medians ($W \approx 2$\,s) with
hysteresis: the system enters the degenerate state when the $f_0$ median
falls below $\tau_{\text{on}}$ and the $\lambda_0$ median below $\tau_2$, and
exits when the $f_0$ median exceeds $\tau_{\text{off}} > \tau_{\text{on}}$ or
the $\lambda_0$ condition fails. The median suppresses single-frame outliers
and the hysteresis band ($\tau_{\text{off}}-\tau_{\text{on}}=0.02$) prevents
chattering ($\approx$1 transition per 1000 frames); evaluating the
statistics on the accumulated map rather than the current scan adds
robustness to instantaneous visibility changes.

When the gate is active, full mitigation ($\gamma=1$) is applied: the
two-condition gate already restricts weighting to genuine absence; a
severity-adaptive blend $\tilde w_m=(1-\gamma)+\gamma\,w_m$ conflates
shallow profiles with dilution, under-powering shallow true degeneracy
(Sec.~\ref{sec:abl}) while still firing in diluted scenes.

\subsection{LiDAR-Only Estimation Limits}\label{sec:limit}
Eq.~\eqref{eq:c1} bounds but does not eliminate along-axis drift: in a
perfectly straight, uniform tunnel the along-axis translation lies in the
geometric null space at every frame, so no reweighting creates absent
information, and over multi-thousand-frame corridors every LiDAR-only method
eventually diverges (full-length GEODE tunnels exceed 700\,m of drift under
KISS-ICP; Sec.~\ref{sec:geode}). Photometry cannot help either:
uniform tunnels carry no along-axis intensity texture, so intensity fusion,
degeneracy-directed intensity weighting, and image-space photometric flow
all leave the geometric null space unresolved. This is the LiDAR-only limit
we report rather than tune around.


\section{Experiments}\label{sec:exp}

\begin{figure*}[t]\centering
\includegraphics[width=\textwidth]{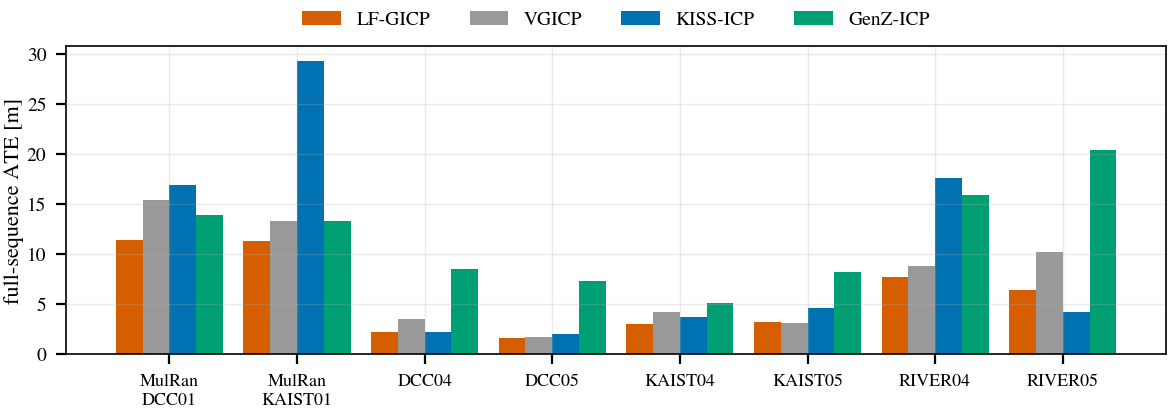}
\caption{Full-sequence ATE on MulRan and HeLiPR
(Table~\ref{tab:ouster} visualized), including the deskewed vanilla VGICP
backend. GenZ-ICP requires 177--185\,GB peak RSS on the four large-area
HeLiPR sequences.}
\label{fig:ousterbars}
\end{figure*}

\subsection{Experimental Setup}\label{sec:setup}
\textbf{Datasets and Sensors.}
We evaluate across platforms spanning over 99,000 frames:
(i)~KITTI Odometry 00--10 (Velodyne HDL-64E),
(ii)~GEODE Urban Tunnel 01--03 and Metro shield tunnel (Velodyne VLP-16,
Livox), (iii)~MulRan DCC01/KAIST01 (Ouster OS1-64), (iv)~HeLiPR DCC04/05,
KAIST04/05, RIVER04/05 (Ouster OS2-128); SubT-MRS underground traces also
appear in the $(f_0,\lambda_0)$ diagnostics of Fig.~\ref{fig:scatter}. All
evaluations are restricted to valid ground-truth spans.

\textbf{Motion Compensation and Baselines.}
Where per-point timestamps exist or are reconstructable (HeLiPR, MulRan),
all methods deskew with CV predictions; KITTI, GEODE, and SubT lack per-point
times and are evaluated without deskewing throughout. We
compare head-to-head against public KISS-ICP and GenZ-ICP, plus vanilla
VGICP odometry (small\_gicp~\cite{smallgicp}; our backend without
degeneracy handling), on identical scans, as pure frame-to-map odometry (no
loop closure, no IMU). On KITTI this means the same raw Velodyne binaries
for every method---including the KISS-ICP and GenZ-ICP numbers in
Table~\ref{tab:kitti}---so rankings reflect the registration front-end, not
dataloader differences.\footnote{Official KISS-ICP reports
$\approx$0.50\% with Velodyne mounting-angle correction
(\texttt{correct\_kitti\_scan}). We omit that correction for
\emph{all} methods (re-run KISS-ICP: 1.053\%); ``lowest'' claims
refer to this matched protocol, not the published 0.50\%.}

\textbf{Metrics and Implementation.}
We report the official KITTI relative translation error (\%), ATE RMSE
after Umeyama alignment~\cite{umeyama}, and segment relative drift (\%)
over 100--800\,m; the pipeline runs on a deterministic C++ core (Eigen,
OpenMP, nanoflann) with a Python front-end.

\textbf{Parameter Calibration.}\label{sec:params}
All gating constants derive via fixed rules from exactly two 500-frame
calibration traces---one well-conditioned (KITTI 00), one degenerate (GEODE
Tunnel01): $\tau_{\text{on}} = 0.165$ is the midpoint of their $f_0$
medians, $\tau_{\text{off}} = \tau_{\text{on}} + 0.02$ enforces the
hysteresis band, $\tau_2 = 0.0143$ is the log-scale geometric mean of their
$\lambda_0$ medians, and $W$ spans 2\,s. Wide sensitivity plateaus around
these values are verified in Sec.~\ref{sec:abl}.

\subsection{KITTI Full-Sequence Relative Error}\label{sec:kitti}
As shown in Table~\ref{tab:kitti}, under the shared raw-scan protocol of
Sec.~\ref{sec:setup} LF-GICP achieves the lowest relative translation error
on 9 of 11 sequences and improves on its vanilla VGICP backend by 28\%.
LF-GICP, KISS-ICP, and GenZ-ICP show comparable ATE (3.98/3.91/3.45\,m) due
to global drift inherent to pure odometry; the relative metric isolates
local tracking, where our formulation excels.

\begin{table}[t]\centering\scriptsize
\caption{KITTI full-sequence relative translation error (\%) under a single
configuration and identical raw scans (best in \textbf{bold}): 9/11 wins,
$-$17.9\%/$-$16.3\% average error vs.\ re-run KISS-ICP/GenZ-ICP, and
$-$28\% vs.\ the vanilla VGICP backend~\cite{smallgicp} without degeneracy
handling. The $\lambda_0$ condition prevents false triggers on open roads
($\le$8\% of frames). See Sec.~\ref{sec:setup} for the KITTI protocol note.}
\label{tab:kitti}
\begin{tabular}{lrcccc}
\toprule
Seq & frames & \textbf{LF-GICP} & VGICP & KISS-ICP & GenZ-ICP\\
\midrule
00 & 4541 & \textbf{0.692} & 1.062 & 0.906 & 0.924\\
01 & 1101 & \textbf{1.822} & 2.390 & 2.120 & 2.238\\
02 & 4661 & \textbf{1.084} & 1.429 & 1.227 & 1.133\\
03 &  801 & 1.207 & 1.165 & 1.071 & \textbf{1.050}\\
04 &  271 & \textbf{0.814} & 1.065 & 0.928 & 0.935\\
05 & 2761 & \textbf{0.577} & 0.841 & 0.778 & 0.666\\
06 & 1101 & \textbf{0.546} & 0.698 & 0.642 & 0.666\\
07 & 1101 & 0.496 & 0.596 & \textbf{0.495} & 0.522\\
08 & 4071 & \textbf{0.878} & 1.110 & 1.112 & 1.035\\
09 & 1591 & \textbf{0.566} & 1.268 & 0.899 & 0.984\\
10 & 1201 & \textbf{0.837} & 1.604 & 1.404 & 1.214\\
\midrule
\textbf{avg} & & \textbf{0.865} & 1.203 & 1.053 & 1.033\\
\bottomrule
\end{tabular}
\end{table}

\subsection{Cross-Sensor Ouster Benchmarks}\label{sec:ouster}
\begin{figure*}[t]\centering
\includegraphics[width=\textwidth]{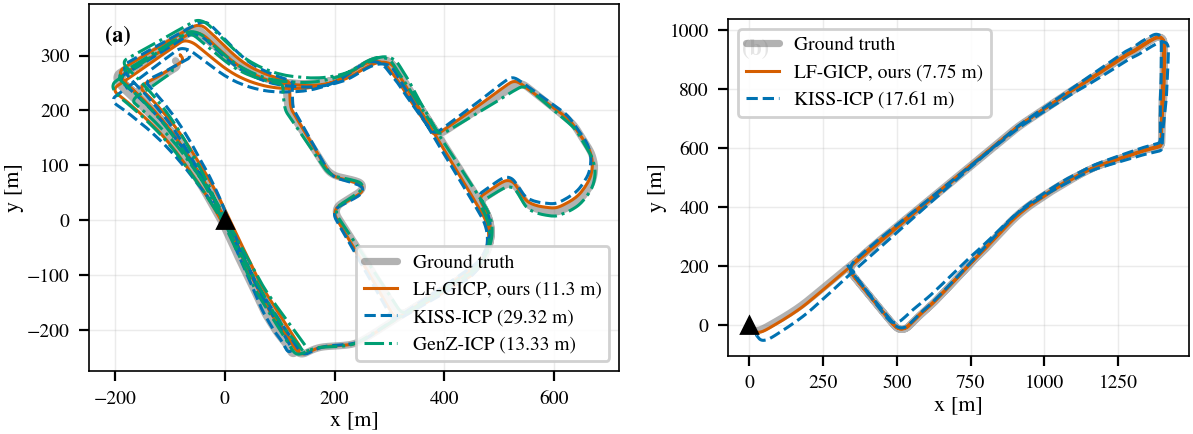}
\caption{Full-sequence trajectories, single frozen configuration
(Umeyama-aligned; ATE in parentheses; $\blacktriangle$ = start).
(a)~MulRan KAIST01 (8142 frames, OS1-64): KISS-ICP accumulates heading
drift over repeated campus loops. (b)~HeLiPR RIVER04 (6114 frames, OS2-128);
GenZ-ICP (15.9\,m) omitted for clarity.}
\label{fig:ouster}
\end{figure*}

Table~\ref{tab:ouster} reports full-sequence ATE on MulRan and HeLiPR under
the single frozen configuration. LF-GICP is best on both MulRan sequences
and achieves the lowest HeLiPR dataset mean (4.04\,m vs.\ 5.28/5.72/10.90\,m
for VGICP/KISS-ICP/GenZ-ICP), beating GenZ-ICP on all six sequences
(Fig.~\ref{fig:ousterbars}); Fig.~\ref{fig:ouster} shows representative
trajectories. The deskewed vanilla backend stays competitive on the
well-conditioned campus routes (3\% ahead on KAIST05)---the gate is largely
silent in these dilution scenes---but trails on MulRan and the riverside
routes. GenZ-ICP's unbounded map also drives peak memory to 177--185\,GB on
the large-area sequences. Segment drift stays comparable to KISS-ICP
(0.77--2.16\% vs.\ 0.69--2.11\%): the ATE gain stems from long-term
heading-drift robustness, not short-term tracking.

\begin{table}[t]\centering\scriptsize
\caption{Full-sequence ATE (m) on MulRan (OS1-64) and HeLiPR (OS2-128),
single frozen motion-compensated configuration (best in \textbf{bold}).
VGICP is the deskewed vanilla backend without degeneracy handling.
$\dagger$runner without deskewing; the four large-area HeLiPR sequences
additionally require 177--185\,GB peak RSS due to unbounded map growth.}
\label{tab:ouster}
\setlength{\tabcolsep}{3.5pt}%
\begin{tabular}{@{}llcccc@{}}
\toprule
 & Seq (frames) & \textbf{LF-GICP} & VGICP & KISS-ICP & GenZ$^\dagger$\\
\midrule
\multirow{2}{*}{MulRan}
 & DCC01 (5409) & \textbf{11.45} & 15.43 & 16.87 & 13.92\\
 & KAIST01 (8142) & \textbf{11.30} & 13.32 & 29.32 & 13.33\\
\midrule
\multirow{6}{*}{HeLiPR}
 & DCC04 (7857) & 2.26 & 3.49 & \textbf{2.19} & 8.51\\
 & DCC05 (10810) & \textbf{1.59} & 1.75 & 2.03 & 7.33\\
 & KAIST04 (12613) & \textbf{2.97} & 4.22 & 3.72 & 5.07\\
 & KAIST05 (12477) & 3.20 & \textbf{3.10} & 4.57 & 8.22\\
 & RIVER04 (6114) & \textbf{7.75} & 8.86 & 17.61 & 15.87\\
 & RIVER05 (7249) & 6.46 & 10.25 & \textbf{4.20} & 20.43\\
\midrule
 & HeLiPR mean & \textbf{4.04} & 5.28 & 5.72 & 10.90\\
\bottomrule
\end{tabular}
\end{table}

\subsection{Degeneracy Handling in Urban Tunnels}\label{sec:geode}
Table~\ref{tab:geode} reports ATE on the GEODE urban tunnels at a 500-frame
horizon. Under the identical frozen configuration, the gate automatically
activates full Fisher-information weighting inside the tunnels ($\lambda_0$
drops to 0.005--0.006) without any environment flag, and LF-GICP beats
KISS-ICP and GenZ-ICP on all three sequences; disabling the gate degrades
tracking to 1.06/2.24/7.54\,m. Vanilla VGICP still tracks at this short
horizon (edging out LF-GICP on mildly degenerate Tunnel03) but diverges by
182--502\,m at full length, where KISS-ICP also drifts $>$700\,m.

\begin{table}[t]\centering\scriptsize
\caption{GEODE Urban Tunnel ATE (m) at a 500-frame horizon under a single
configuration (best in \textbf{bold}). $\ddagger$vanilla VGICP diverges by
182--502\,m over the full sequences.}
\label{tab:geode}
\begin{tabular}{lcccc}
\toprule
Sequence & \textbf{LF-GICP} & VGICP$^\ddagger$ & KISS-ICP & GenZ-ICP\\
\midrule
Urban\_Tunnel01 & \textbf{0.83} & 0.99 & 1.91 & 1.36\\
Urban\_Tunnel02 & \textbf{1.93} & 2.66 & 13.70 & 10.90\\
Urban\_Tunnel03 & 5.20 & \textbf{3.65} & 5.45 & 5.37\\
\bottomrule
\end{tabular}
\end{table}



\subsection{Ablation Studies}\label{sec:abl}
\textbf{Detection Signal.} Replacing the field $\M$ \eqref{eq:M} with the
GN-Hessian translation-block fraction as the gate input fails entirely on
GEODE: tunnel and open driving become indistinguishable
(Sec.~\ref{sec:hidden}) and mitigation never activates.
Table~\ref{tab:creg} confirms the masking is intrinsic: across
$\beta\in\{0.1,0.5,1.0,2.0\}$ the Hessian fraction stays flat
($\approx0.32$) while $f_0$ continues to expose the tunnel.

\begin{table}[t]\centering\footnotesize
\caption{Regularization sweep: the Hessian fraction stays flat while $f_0$
exposes the tunnel degeneracy throughout.}
\label{tab:creg}
\begin{tabular}{lcccc}
\toprule
Regularization $\beta$ & 0.1 & 0.5 & 1.0 & 2.0\\
\midrule
Tunnel03 Hessian $\lambda_{\min}/\mathrm{tr}$ & 0.316 & 0.314 & 0.316 & 0.324\\
Tunnel03 $f_0$ (Ours) & 0.157 & 0.138 & 0.128 & 0.125\\
KITTI seq00 ATE (m) & 0.401 & 0.374 & 0.349 & 0.318\\
\bottomrule
\end{tabular}
\end{table}

\textbf{Absence Condition.} An $f_0$-only gate fires mistakenly on
anisotropic yet dense scenes, degrading KITTI from 0.865\% to 0.922\% and
HeLiPR campus errors by up to 1.4$\times$; adding the $\lambda_0 < \tau_2$
condition deactivates mitigation there while leaving genuine absence domains
(GEODE) unchanged; disabling the gate entirely degrades all GEODE
sequences to 1.06/2.24/7.54\,m (Sec.~\ref{sec:geode}).

\textbf{Mitigation Strength.} Under the same gate, the severity-adaptive
blend $\tilde w_m=(1-\gamma)+\gamma\,w_m$ with
$\gamma=\mathrm{clamp}((\tau_{\text{on}}-f_0)/\tau_{\text{on}},0,1)$
under-powers mild degeneracy on GEODE Tunnel03 (ATE $6.83$ vs.\
$5.20$\,m) while only marginally recovering KITTI ($\approx$0.88\%): any
$f_0$-monotone strength rule that spares diluted scenes is too weak for
shallow true degeneracy; the binary absence gate with full weighting
($\gamma{=}1$) resolves the tension without new tuning parameters
(Table~\ref{tab:strength}).

\begin{table}[t]\centering\footnotesize
\caption{Mitigation strength ablation on KITTI (avg.\ relative error) and
GEODE Tunnel03. Best results are in \textbf{bold}.}
\label{tab:strength}
\setlength{\tabcolsep}{3.5pt}
\begin{tabular}{@{}lcc@{}}
\toprule
Strategy & KITTI (\%) & Tunnel03 (m)\\
\midrule
$\gamma{=}1$, $\lambda_0$ gate (Ours) & \textbf{0.865} & \textbf{5.20} \\
$\gamma{=}1$, $f_0$-only gate & 0.922 & \textbf{5.20} \\
Blend ($\gamma{<}1$), $f_0$-only & $\approx$0.88 & 6.83 \\
\bottomrule
\end{tabular}
\end{table}

\textbf{Parameter Sensitivity.} Independent sweeps confirm wide plateaus:
accuracy is invariant for $\tau_{\text{on}} \in [0.15, 0.19]$ and
$W \in [5, 40]$, and $\tau_2$ separates its classes with a clear margin
(tightest gap: 0.013 vs.\ 0.017)---stable calibration outputs, not tuning
knobs.



\section{Conclusion}\label{sec:conclusion}
We detect LiDAR degeneracy from a voxel-normal localizability field
invisible to the regularization-floored GN Hessian, gating
Fisher-information mitigation on $f_0$ and $\lambda_0$ and eliminating
per-environment parameters. One frozen two-trace calibration yields the
lowest KITTI relative error under the matched protocol of
Sec.~\ref{sec:setup}, beats KISS-ICP and GenZ-ICP on all GEODE tunnels,
and leads MulRan and the HeLiPR mean across four sensor types; straight
uniform tunnels remain unobservable to any LiDAR-only method.

\bibliographystyle{IEEEtran}
\bibliography{paper_final}

\clearpage
\setcounter{section}{0}
\setcounter{subsection}{0}
\setcounter{table}{0}
\setcounter{figure}{0}
\setcounter{algorithm}{0}
\renewcommand{\thesection}{S\Roman{section}}
\renewcommand{\thetable}{S\Roman{table}}
\renewcommand{\thefigure}{S\arabic{figure}}
\renewcommand{\thealgorithm}{S\arabic{algorithm}}

\begin{center}
{\LARGE\bf Supplementary Material for\\
LF-GICP: Parameter-Free Degeneracy-Aware LiDAR Odometry\\
via a Voxel-Normal Localizability Field}
\end{center}
\vspace{1em}

\section{Derivations}\label{sec:supp-deriv}

This section derives the analytic statements of the main-paper Method
section (Sec.~\ref{sec:method}) that are given there without proof,
in the order they appear in the paper.

\subsection{Closed-Form Constant-Velocity Deskewing}
\label{sec:deriv-deskew}
We derive the per-point correction of Eq.~\eqref{eq:deskew}. Let
$\bm\xi=\log(\bm T_{k-2}^{-1}\bm T_{k-1})\in\sethree$ be the body-frame
twist accumulated over the previous scan period. Under the
constant-velocity assumption the sensor continues to move with the same
twist during the current scan, so the sensor pose at normalized
in-scan time $\tau\in[0,1]$ is the geodesic (screw) interpolation
\begin{equation}
\bm T(\tau)=\bm T_{\text{end}}\exp\!\big((\tau-1)\,\bm\xi\big),
\label{eq:screw}
\end{equation}
where $\bm T_{\text{end}}=\bm T(1)$ is the scan-end pose in which the
registration cost \eqref{eq:cost} is expressed. A point $\bm p_i$
measured at time $\tau_i$ lives in the frame $\bm T(\tau_i)$; its
coordinates in the scan-end frame are
$\bm T_{\text{end}}^{-1}\bm T(\tau_i)\,\bm p_i
=\exp((\tau_i-1)\bm\xi)\,\bm p_i$, which is
Eq.~\eqref{eq:deskew}. The exponential shares one axis--angle pair
across the scan, so the map can be evaluated with a single Rodrigues
expansion batched over all points (one closed-form
$\mathrm{SE}(3)$ pass, no per-point iteration).

\subsection{The Isotropic Floor of the Regularized Hessian}
\label{sec:deriv-floor}
We derive Eq.~\eqref{eq:floor} and the resulting scene-independent
eigenvalue ratio. A planar voxel has raw covariance
$\bm C_j=\sigma_t^2(\bm I-\n_j\n_j^\top)+\sigma_n^2\,\n_j\n_j^\top$
with $\sigma_n^2\ll\sigma_t^2$. Since $\bm I-\n_j\n_j^\top$ and
$\n_j\n_j^\top$ are complementary orthogonal projectors, the
regularization $\bm C_j+\beta\bm I$ acts on each eigenspace
separately, and inversion acts per eigenspace as well:
\begin{equation}
\Om_j=(\bm C_j+\beta\bm I)^{-1}
=\kappa_t(\bm I-\n_j\n_j^\top)+\kappa_n\,\n_j\n_j^\top,
\label{eq:invsplit}
\end{equation}
with $\kappa_t=1/(\sigma_t^2+\beta)$ and
$\kappa_n=1/(\sigma_n^2+\beta)$. Writing
$\kappa_t(\bm I-\n\n^\top)+\kappa_n\n\n^\top
=\kappa_t\bm I+(\kappa_n-\kappa_t)\n\n^\top$ and summing over
correspondences gives Eq.~\eqref{eq:floor}:
$\Hess_{\rho\rho}=F\bm I+\bm S$ with
$F=\sum_m\kappa_t^{(m)}$ and
$\bm S=\sum_m(\kappa_n^{(m)}-\kappa_t^{(m)})\n_m\n_m^\top\succeq0$.

Let $0\le s_1\le s_3$ be the extreme eigenvalues of $\bm S$. Then
\begin{equation}
\frac{\lambda_{\min}(\Hess_{\rho\rho})}{\lambda_{\max}(\Hess_{\rho\rho})}
=\frac{F+s_1}{F+s_3}
\;\ge\;\frac{F}{F+\mathrm{tr}(\bm S)}
\;\ge\;\frac{\bar\kappa_t}{\bar\kappa_n},
\label{eq:ratiobound}
\end{equation}
where bars denote per-correspondence averages and the last bound is the
worst case of all normals concentrated on one axis. Both $F$ and
$\bm S$ scale linearly with the number of correspondences, so the ratio
converges to a constant determined only by the material statistics
$(\sigma_n^2,\sigma_t^2,\beta)$---not by the scene geometry. For
representative planar-voxel values
($\sigma_n^2\!\approx\!10^{-2}$, $\sigma_t^2\!\approx\!1$,
$\beta\!=\!1$) the bound is $\approx0.5$, consistent with the tunnel
ratios measured in the main paper ($\lambda_{\min}/\lambda_{\max}=0.87$,
$\lambda_{\min}/\mathrm{tr}=0.31$).
As $\beta\to0$, $\kappa_t\to1/\sigma_t^2$ stays finite: the floor is
intrinsic to the full-covariance formulation, and exposing the
degeneracy would additionally require $\sigma_n^2\to0$, which sensor
noise and surface roughness prevent. This is why the measured ratio
stays flat ($\approx0.32$) across the entire $\beta$ sweep of
main-paper Table~\ref{tab:creg}.

\subsection{Properties of the Localizability Field}
\label{sec:deriv-field}
\textbf{Planarity spectrum.} For the planar information matrix
\eqref{eq:invsplit} the eigenvalues are $[\kappa_t,\kappa_t,\kappa_n]$,
so $\rho_v=(\lambda_3-\lambda_2)/\lambda_3=(\kappa_n-\kappa_t)/\kappa_n$,
which approaches $1$ for an ideal plane
($\sigma_n^2\to0$, hence $\kappa_n\to\infty$ at $\beta=0$) and $0$ for
an isotropic voxel ($\kappa_n=\kappa_t$).

\textbf{Range of $f_0$.} $\M$ in Eq.~\eqref{eq:M} is a sum of
positive-semidefinite rank-one terms, so
$0\le\lambda_{\min}(\M)\le\tfrac13\mathrm{tr}(\M)$, giving
$f_0\in[0,\tfrac13]$ with the maximum attained only for an isotropic
normal distribution.

\textbf{Relation to the unregularized FIM.} If every contributing voxel
is an ideal plane ($\rho_v=1$) with common normal information
$\kappa_n$, the unregularized translation FIM is
\begin{equation}
\sum_v\Om_v^{\,\beta=0}
=\sum_v\kappa_n\n_v\n_v^\top+\mathcal O(\kappa_t)
=\kappa_n\M+\mathcal O(\kappa_t),
\end{equation}
so $\mathrm{null}(\M)$ coincides with the unconstrained translation
subspace. Moreover, adding any isotropic term $c\,\bm I$ (the floor of
Eq.~\eqref{eq:floor}) shifts all eigenvalues of $\Hess_{\rho\rho}$
equally but leaves its eigenvectors---and $\M$ itself, which is built
before regularization---unchanged.

\subsection{Absence vs.\ Dilution: the $\epsilon$-Model}
\label{sec:deriv-eps}
We derive the approximations
$f_0\approx\epsilon/(1+\epsilon)$ and $\lambda_0\approx\epsilon\bar\rho$
quoted in Sec.~\ref{sec:lam0}. Consider $N_\perp$ planar voxels whose
normals lie uniformly in the cross-section plane orthogonal to the weak
axis $\bm u$, plus $N_\parallel=\epsilon N_\perp$ voxels with normals
along $\bm u$, all with mean planarity $\bar\rho$. By symmetry the
in-plane normals distribute their mass equally over the two
cross-section axes:
\begin{equation}
\M\approx\bar\rho\Big[\epsilon N_\perp\,\bm u\bm u^\top
+\tfrac{N_\perp}{2}(\bm I-\bm u\bm u^\top)\Big],
\end{equation}
with eigenvalues $\bar\rho\,\epsilon N_\perp$ (along $\bm u$) and
$\bar\rho N_\perp/2$ (twice). For $\epsilon<\tfrac12$ the weak axis is
$\bm u$ and, with $|\mathcal V|=(1+\epsilon)N_\perp$,
\begin{equation}
f_0=\frac{\epsilon}{1+\epsilon},\qquad
\lambda_0=\frac{\bar\rho\,\epsilon}{1+\epsilon}\approx\epsilon\bar\rho .
\end{equation}
The fraction $f_0$ cancels both $\bar\rho$ and the voxel count---it is
scale-free and cannot tell whether a low value stems from a genuinely
empty axis (absence, $\epsilon\to0$) or from a scene where facade
normals merely outnumber fronto-facing ones (dilution, small but
nonzero $\epsilon$ with large absolute mass). The per-voxel mass
$\lambda_0$ retains the factor $\bar\rho\,\epsilon$, i.e., the absolute
weak-axis surface mass, which is what separates the two regimes in
main-paper Fig.~\ref{fig:scatter}.

\subsection{Properties of the FIM Weighting}
\label{sec:deriv-weight}
Let $q_m=\vmin^\top\J_m^\top\Om_m\J_m\vmin\ge0$ be the weak-axis
information of correspondence $m$ and
$\bar q=\tfrac1{N_c}\sum_m q_m$ its mean, so that
$w_m=q_m/\bar q$ in Eq.~\eqref{eq:c1}.

\textbf{Scale preservation.} By construction
$\tfrac1{N_c}\sum_m w_m=1$: the mean information weight is unchanged,
so the effective scale of $\sum_m w_m\Om_m$ and the adaptive threshold
statistics that depend on it are preserved.

\textbf{Weak-axis amplification.} After reweighting,
\begin{equation}
\vmin^\top\Hess'\vmin
=\frac{\sum_m q_m^2}{\bar q}
\;\ge\;\sum_m q_m=\vmin^\top\Hess\vmin,
\end{equation}
by the quadratic--arithmetic mean inequality, with equality iff all
$q_m$ are equal. Reweighting therefore never reduces the weak-axis
information and strictly amplifies it exactly when that information is
concentrated in a minority of correspondences---the degenerate case.

\textbf{No virtual information.} Each term of
$\Hess'=\sum_m w_m\J_m^\top\Om_m\J_m$ is a nonnegative multiple of an
existing measurement term, so
$\mathrm{range}(\Hess')\subseteq\mathrm{range}(\Hess)$: directions in
the data null space receive no information from the reweighting, in
contrast to isotropic-floor or prior-locking schemes
(cf.\ Sec.~\ref{sec:supp-abl}).

\section{Complete Parameter Set and Per-Frame Algorithm}\label{sec:supp-params}

This section complements the main-paper Method section
(Sec.~\ref{sec:method} and the calibration protocol of
Sec.~\ref{sec:params}). Table~\ref{tab:params} lists \emph{every}
constant used by LF-GICP, grouped by how it is set: \emph{Auto}
quantities are recomputed from data at every frame and involve no
tuning; \emph{Global} constants are derived once from the two
calibration traces by the fixed rules of the main paper and then frozen
across all datasets, sensors, and platforms; \emph{Sensor} constants
follow the beam-count convention shared with the baselines and are not
degeneracy-related. Algorithm~\ref{alg:full} then gives the complete
per-frame pipeline: deskewing, construction of the localizability field
$\M$ and its two statistics $f_0$ and $\lambda_0$, the temporal-median
hysteresis gate, and the Gauss--Newton loop with Fisher-information
correspondence weighting applied only while the gate is active.

\begin{table}[t]\centering\footnotesize
\caption{Complete parameter set. Auto = data-driven per frame; Global =
one value frozen across all datasets/sensors (derived from the two
calibration traces); Sensor = per sensor class.}
\label{tab:params}
\setlength{\tabcolsep}{3pt}
\begin{tabular}{@{}lll@{}}
\toprule
Class & Parameter & Value / rule\\
\midrule
\multirow{3}{*}{\textbf{Auto}}
& mitigation on/off & $\mathrm{med}_W(f_0),\mathrm{med}_W(\lambda_0)$ gate, Alg.~\ref{alg:full}\\
& degenerate axis $\bm u$ & null eigenvector of $\M$ \eqref{eq:M}\\
& deskew motion & constant-velocity prediction\\
\midrule
\multirow{5}{*}{\textbf{Global}}
& anisotropy enter $\tau_{\text{on}}$ & $0.165$\\
& anisotropy exit $\tau_{\text{off}}$ & $0.185$ (robust preset: $0.30$)\\
& absence threshold $\tau_2$ & $0.0143$\\
& gate window $W$ & $20$ frames\\
& field subsample $L_{\max}$ & $4096$ voxels\\
\midrule
\multirow{6}{*}{\textbf{Sensor}}
& voxel size & $1.0$\,m ($\ge$64 beams) / $0.5$\,m ($\le$32)\\
& source voxel & $0.3$\,m / $0.25$\,m\\
& max range & $80$\,m\\
& GN iterations & $12$\\
& Huber $\delta$ & $1.0$\\
& map: radius / frames / corr & age-based / $500$ / $2.0$\,m\\
\bottomrule
\end{tabular}
\end{table}

\begin{algorithm}[t]
\caption{LF-GICP per-frame odometry}\label{alg:full}
\begin{algorithmic}[1]
\State \textbf{Input:} scan (+ point times), target voxel map,
       state $(\text{deg},\mathcal W,\mathcal W_\lambda)$
\State deskew scan by constant-velocity motion if point times available
\State $\M\leftarrow\sum_{v}\rho_v\n_v\n_v^\top$ over $\le L_{\max}$ voxels;
       $f_0\leftarrow\lambda_{\min}(\M)/\mathrm{tr}(\M)$;
       $\lambda_0\leftarrow\lambda_{\min}(\M)/|\mathcal V|$
       \Comment{Eqs.~\eqref{eq:M},\,\eqref{eq:lam0}}
\State push to windows (length $W$);
       $\tilde f\leftarrow\mathrm{med}(\mathcal W)$,
       $\tilde\lambda\leftarrow\mathrm{med}(\mathcal W_\lambda)$
\If{$\neg\text{deg}$ \textbf{and} $\tilde f<\tau_{\text{on}}$
    \textbf{and} $\tilde\lambda<\tau_2$} $\text{deg}\leftarrow\text{true}$
\ElsIf{$\text{deg}$ \textbf{and} ($\tilde f>\tau_{\text{off}}$
    \textbf{or} $\tilde\lambda\ge\tau_2$)} $\text{deg}\leftarrow\text{false}$
\EndIf
\For{GN iteration}
  \State build $\Hess,\bm b$ \eqref{eq:gn}; if $\text{deg}$, apply FIM
         weight $w_m$ \eqref{eq:c1} to each $\Om_m$
  \State $\delta\bm\xi\leftarrow$ solve; $\bm T\leftarrow\exp(\delta\bm\xi)\bm T$
\EndFor
\State insert scan into map; \textbf{return} $\bm T$
\end{algorithmic}
\end{algorithm}

\section{Additional Result Visualizations}\label{sec:supp-figs}

The figures in this section visualize results that are reported
numerically in the main paper, following the order of the main-paper
experiment section. Qualitative trajectories for GEODE Tunnel~02, MulRan
KAIST01, and HeLiPR RIVER04 are shown in main-paper
Figs.~\ref{fig:teaser} and~\ref{fig:ouster}.

\begin{figure}[t]\centering
\includegraphics[width=.8\linewidth]{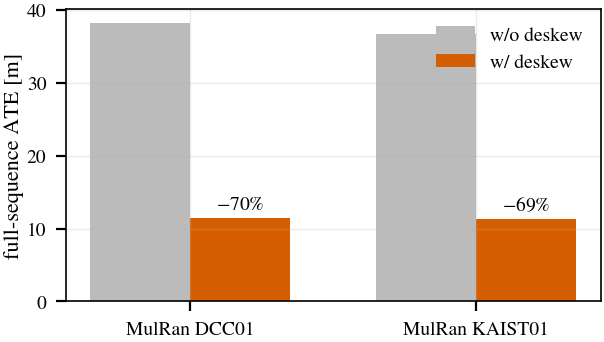}
\caption{Motion compensation on MulRan (azimuth-reconstructed per-point
times): deskewing is the dominant correction for the point-to-distribution
backend, independent of degeneracy handling.}
\label{fig:deskew}
\end{figure}

\textbf{Motion compensation (Fig.~\ref{fig:deskew}).}
Figure~\ref{fig:deskew} supports the motion-compensation protocol of
the experimental setup (Sec.~\ref{sec:setup}): on MulRan, with
per-point times reconstructed from azimuth, constant-velocity deskewing
is the dominant accuracy correction for the point-to-distribution
backend, independent of and orthogonal to the degeneracy handling. This
motivates applying the same deskewing to the strongest baseline (KISS-ICP)
for a fair comparison.

\begin{figure*}[t]\centering
\includegraphics[width=.92\textwidth]{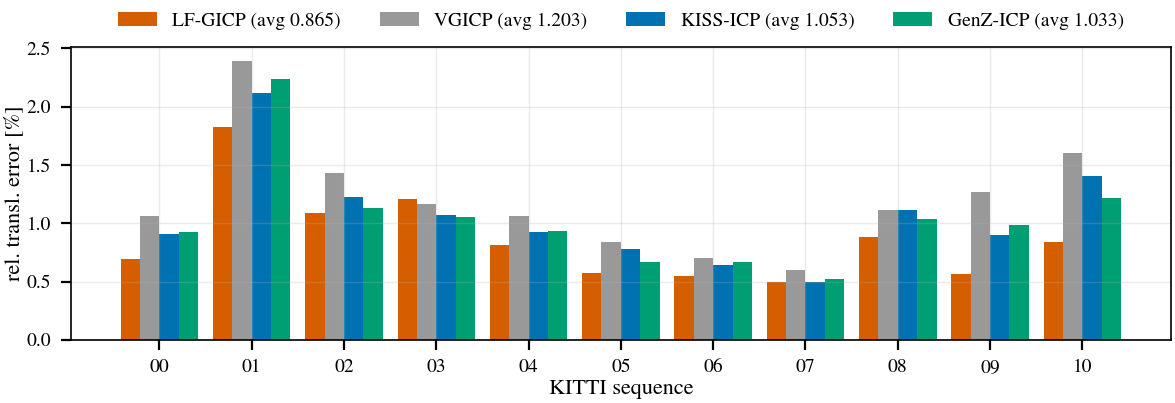}
\caption{KITTI full-sequence relative translation error per sequence
(main-paper Table~\ref{tab:kitti} visualized), including the vanilla
VGICP backend without degeneracy handling.}
\label{fig:kittibars}
\end{figure*}

\textbf{Per-sequence KITTI bars (Fig.~\ref{fig:kittibars}).}
Figure~\ref{fig:kittibars} visualizes the per-sequence KITTI relative
translation errors of main-paper Table~\ref{tab:kitti}, including the
vanilla small\_gicp VGICP backend, showing that the average win is not
driven by a single sequence. The MulRan/HeLiPR bar chart is in the main
paper (Fig.~\ref{fig:ousterbars}).

\section{SubT-MRS: Detection Generalizes; a Representation Limit}
\label{sec:subt}
\begin{figure}[t]\centering
\includegraphics[width=.82\linewidth]{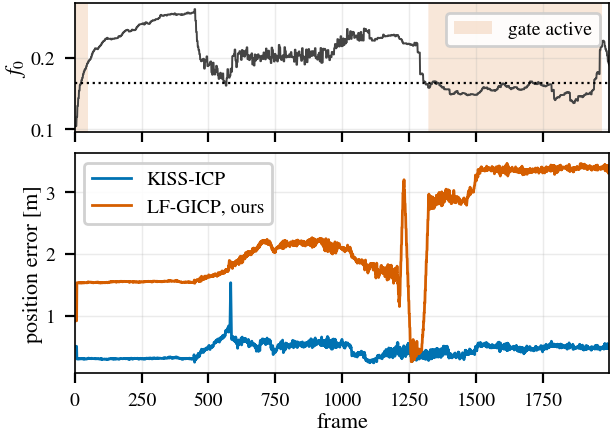}
\caption{SubT rock mine failure case. Top: the gate triggers accurately on
tunnel entry. Bottom: LF-GICP error growth stems from the representation
limit of Gaussian voxel maps, not a detection failure.}
\label{fig:mine}
\end{figure}

This section reports the SubT-MRS study referenced in the main-paper
experimental setup (Sec.~\ref{sec:setup}); the corresponding
gate/error timeline is shown in Fig.~\ref{fig:mine}.
On six SubT-MRS underground sequences (2000-frame horizons) LF-GICP and
KISS-ICP split 3--3 (LF wins Urban\_UGV2 $0.27$ vs $0.40$, Final\_UGV2
$0.14$ vs $0.20$, Final\_UGV3 $0.08$ vs $0.40$), while GenZ-ICP with
SubT-tuned parameters is best on all six---its adaptive
point-to-plane/point blend suits confined spaces. The two clear LF
losses share one property: rough \emph{natural} surfaces (rock mine
Final\_UGV1 $2.32$ vs KISS $0.45$; Laurel cave $1.28$ vs $0.59$). The
failure is forensically \emph{not} a detection failure: the gate fires
correctly (cave $\lambda_0{=}0.009$, gate active $99\%$), local accuracy
equals KISS (RPE $0.11$ for both), and the drift is along-axis slip that
survives every configuration change we tested (gate off, map policy,
range, finer voxels---$0.25$\,m voxels are catastrophic at $12.3$\,m---%
correspondence radius, motion threshold) \emph{and} the null-axis
information floor of Sec.~\ref{sec:supp-abl}. The per-voxel Gaussian
compression of point-to-distribution registration destroys along-axis
micro-texture that point-based maps retain; the same argument the
intensity study (Sec.~\ref{sec:int}) makes for photometry applies to
the map representation itself. Degeneracy-aware weighting is necessary
but cannot substitute for a representation that retains the signal---we
state this as the honest boundary of the backend, orthogonal to the
detection/gating contribution.

\section{Intensity Cannot Recover the Axis}\label{sec:int}
This section details the photometric null-space study summarized in the
main paper (Sec.~\ref{sec:limit}).
On the GEODE tunnel freeze region we test three photometric variants.
(i)~Uniform intensity fusion (4D geo-photometric residual): erratic, no
fix ($f_0$-region ATE across $\alpha\!\in\![0,2]$ non-monotone,
$>\!50$\,m at full length). (ii)~Degeneracy-directed per-voxel intensity
(weight $\propto(\nabla I\!\cdot\!\bm u)^2$): \emph{zero} effect on ATE
across all gains (T03@1000 flat at $\approx108$\,m). (iii)~Image-space
photometric flow from the organized 16-ring scan: inter-frame azimuth
shift uncorrelated with ground-truth along-axis motion (image
autocorrelation $\approx0.8$, near-invariant). All three confirm the
uniform tunnel has no along-axis intensity texture, matching the
geometric null space.

\section{Additional Ablation Detail}\label{sec:supp-abl}

The quantitative ablations of the main paper
(Sec.~\ref{sec:abl})---Detection Signal (Table~\ref{tab:creg}),
Absence Condition and Mitigation Strength (Table~\ref{tab:strength})---%
are reported there in full. This section records rejected alternatives
and the soft-vs.-hard constraint comparison.

\textbf{Rejected Alternatives.} Three architectural variants were
evaluated and rejected: (i) a statistical Mahalanobis auto-trim, which
degrades KITTI and GEODE accuracy by 12--13\%, (ii) an isotropic
information floor, which worsens SubT mine errors by filling the null
space with virtual constraints that overwrite raw geometric data, and
(iii) gating via per-correspondence weak-axis information concentration,
which fails to separate the help/hurt classes due to identical
top-decile weight distributions (0.430) across both regimes.

\textbf{Soft Weighting vs.\ Hard Constraints.} Replacing our soft FIM
reweighting with a hard equality constraint that locks the degenerate
axis to a constant-velocity prior (similar to X-ICP) consistently
degrades performance, causing GEODE Urban to diverge and worsening
GEODE Metro tracking. Because the regularized solver retains a weak but
valid structural along-axis signal, hard-locking discards this geometric
information in favor of an unreliable dead-reckoning prior. Soft
weighting stabilizes the linear system while preserving all remaining
physical cues.

\section{Speed}\label{sec:speed}
Field computation is one stride-subsampled ($\le L_{\max}$) closed-form
$3\times3$ eigen-pass per frame plus one $3\times3$ eigendecomposition
of $\M$ (microseconds, independent of map size); deskewing is a single
batched closed-form $\mathrm{SE}(3)$ pass over the scan. Outside
degenerate stretches the gate runs plain GICP, so the degeneracy
machinery itself is effectively free. End-to-end throughput is
sensor-dependent: $\sim$15--20\,Hz on 64-beam scans (KITTI, MulRan) and
$\sim$5\,Hz on full-density 128-beam HeLiPR scans ($\sim$10$^5$
pts/frame) with our unoptimized front-end---about $2\times$ slower than
KISS-ICP overall; the contribution of this paper is accuracy and
automation, not speed.

\section{Scope and Honesty}\label{sec:honest}
``Parameter-free'' means \emph{no per-environment parameter}: the
degeneracy decision is data-driven from the field statistics and the
mitigation applies full FIM weighting (no strength to tune); the
remaining constants (Table~\ref{tab:params}) are single global values
derived from two calibration traces and frozen, with measured
sensitivity plateaus. Best-method claims are: KITTI relative error
(full sequences, head-to-head), GEODE tunnels at the 500-frame horizon
vs.\ KISS-ICP and GenZ-ICP, MulRan full-sequence ATE, and the HeLiPR
dataset mean.
Not claimed: KITTI ATE (comparable, drift-dominated); HeLiPR DCC04
($3\%$ behind deskewed KISS), KAIST05 ($3\%$ behind the deskewed vanilla
VGICP backend, $3.20$ vs.\ $3.10$\,m), and RIVER05; segment relative
drift on MulRan/HeLiPR (within $0.1$--$0.2$\,pp of KISS, split); GEODE
Tunnel03 at the 500-frame horizon vs.\ vanilla VGICP ($3.65$ vs.\
$5.20$\,m), which nevertheless diverges by $182$--$502$\,m at full
length; SubT rough natural surfaces (Sec.~\ref{sec:subt}); full-length
straight tunnels (LiDAR-only limit, all methods fail). GenZ-ICP
comparisons carry a protocol note (its runner lacks deskewing);
completing the four large-area HeLiPR sequences required a 224\,GB
machine (177--185\,GB peak RSS from unbounded map growth).

\end{document}